# UNLOCKING THE POTENTIAL OF COLLABORATIVE AI – ON THE SOCIO-TECHNICAL CHALLENGES OF FEDERATED MACHINE LEARNING

*Research Paper*

Tobias Müller, Technical University of Munich, School of Computation, Information and Technology, Department of Computer Science, Germany and SAP SE, Germany, tobias1.mueller@tum.de

Milena Zahn, Technical University of Munich, School of Computation, Information and Technology, Department of Computer Science, Germany and SAP SE, Germany, milena.zahn@tum.de

Florian Matthes, Technical University of Munich, School of Computation, Information and Technology, Department of Computer Science, Germany, matthes@tum.de

## Abstract

*The disruptive potential of AI systems roots in the emergence of big data. Yet, a significant portion is scattered and locked in data silos, leaving its potential untapped. Federated Machine Learning is a novel AI paradigm enabling the creation of AI models from decentralized, potentially siloed data. Hence, Federated Machine Learning could technically open data silos and therefore unlock economic potential. However, this requires collaboration between multiple parties owning data silos. Setting up collaborative business models is complex and often a reason for failure. Current literature lacks guidelines on which aspects must be considered to successfully realize collaborative AI projects. This research investigates the challenges of prevailing collaborative business models and distinct aspects of Federated Machine Learning. Through a systematic literature review, focus group, and expert interviews, we provide a systemized collection of socio-technical challenges and an extended Business Model Canvas for the initial viability assessment of collaborative AI projects.*

Keywords: Federated Machine Learning, Collaborative Data Processing, Business Model, Alliances

## 1  Introduction

Artificial Intelligence (AI) had an immense economic impact in the last couple of years. In 2021 alone, the market of AI-based services including software, hardware and services exceeded 500$ billion with a five-year compound annual growth rate of 17.5% (Forradellas and Gallastegui, 2021). The potential profitability raise is currently estimated by an average of 38%, which implies an economic impact of $14 trillion until 2035[1]. Unmistakably, the usage of AI enables new, unprecedented business models with a monumental impact on the industry. The main enabler for this disruptive new market is the emergence of big data, which forms the fundamental basis for AI systems. Even though vast amounts of data is freely available, a considerable amount of the world's data is scattered, stored and locked up in decentralized IoT devices and data silos. Naturally, the siloed data is hardly accessible, leaving a large portion of already generated data, and therefore economic potential, largely untapped. The emergence of data silos is strengthened by data protection laws and regulations such as the General Data Protection Regulation (GDPR), California Consumer Privacy Act, Cyber Security Law and the General Principles of the Civil Law. These regulations justifiably aim to protect the privacy of individuals and therefore restrict direct data sharing between different parties (Li et al., 2022). This protection of privacy is an important pursuit but leads to more data silos and therefore unused economic potential.

---

[1] https://www.accenture.com/fr-fr/_acnmedia/36dc7f76eab444cab6a7f44017cc3997.pdf

Federated Machine Learning (FedML) introduced by McMahan et al. (2016) is a novel machine learning (ML) technology with the potential of building prediction models of decentralized and therefore siloed datasets. In contrast to traditional, centralized ML, FedML systems initially train a global ML model which is then distributed to all participants. Then, each participant individually trains the model locally on their own dataset. The clients solely return the update gradient resulting from the local training. Through this model-to-data approach, the data never leaves the client's device, but still enables the development of a joint ML model. Thus, FedML enables tapping the potential of big data without privacy leakage.

FedML technically has the potential to leverage siloed data while still preserving the intellectual property (IP) and privacy of each individuals' dataset. Hence, FedML enables the usage of currently untapped data and therefore brings the potential to be the catalyst for novel, disruptive business model innovation and locking unprecedented value from siloed data. However, this requires the collaboration of multiple parties which own these data silos. Hence, a collaborative business model is needed as a framework for how value can be created, and different parties can be incentivized for participating in such a collaborative network. Setting up collaborative business models is complex and a potential reason for failure. The current literature lacks guidelines for decision-makers on which aspects must be considered for the successful realization of collaborative AI projects.

This work aims towards closing this knowledge gap. More specifically, we investigate the challenges of prevailing collaborative business models through a systematic literature review and identify distinct aspects of collaborative FedML projects by conducting a focus group interview and multiple expert interviews. We work towards a systemized collection of socio-technical challenges and an easily consumable business model canvas (BMC) to aid decision-makers in the initial viability assessment of collaborative AI projects. Summarized, we aim to answer the following research questions (RQs):

**RQ1:** What are the general challenges of collaborative business models?

**RQ2:** What are the aspects of inter-organizational FedML business models in relation to prevailing collaborative business models?

**RQ3:** Which aspects and attributes should be considered for inter-organizational FedML projects and how can these be structured into an extended BMC?

To address these research questions, we first describe the theoretical background of our study by introducing Federated Machine Learning and providing background information on collaborative business models (section 2). Following, we elaborate on our tripartite research methodology, which consists of a systematic literature review, in-depth focus group interviews, and semi-structured expert interviews (section 3). Subsequently, we present the results of our research including a systemized overview of challenges for collaborative business models, a structured list of distinct socio-technical aspects for FedML projects and a proposal for a corresponding extended BMC (section 4). Finally, we discuss our work by reflecting the underlying research problem and research gaps. The discussion is followed by a summary of our contributions, answers to the RQs and limitations of our work. Our study concludes with an outline of future research (section 5).

## 2 Theoretical Background

The following section presents the theoretical background of our study. We first describe the motivation, terminologies, and the basic concept of FedML as originally proposed by McMahan et al. (2016). Subsequently, we provide general background information on business models to establish a common understanding for this study. Finally, we elaborate on collaborative business models and corresponding extensions of the BMC by Osterwalder and Pigneur (2010).

## 2.1 Federated Machine Learning

A classic ML approach requires the collaborating participants to assemble their datasets in a central location and train a unique ML model $M_{SUM}$, exposing the data to each other and the central server. The participants thereby risk losing their data sovereignty and IP, which inhibits companies to collaborate and share data (Schomakers et al., 2020). Introduced by McMahan et al. (2016), FedML counteracts the need of sharing datasets through a model-to-data approach. As illustrated in Figure 1, a global ML model is chosen, which is distributed amongst all clients. The clients train the model locally on their individual dataset. The update gradients are sent back to the server and used to improve the global model. Thereby, FedML enables data owners to train a joint model $M_{FED}$ without the need to disclose their data.

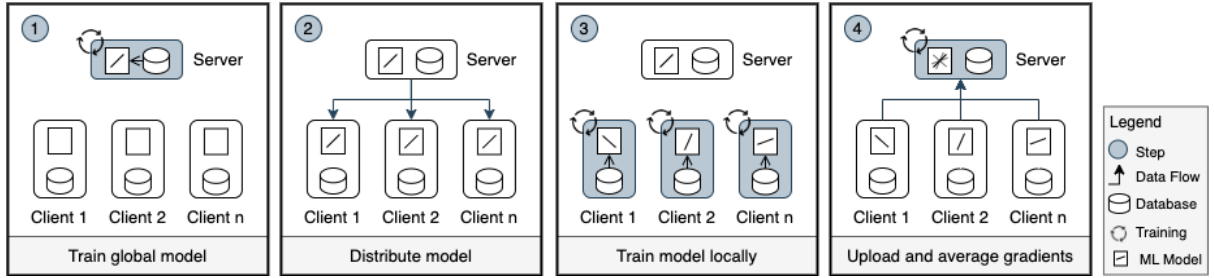

*Figure 1.  One iteration of the Federated Machine Learning process (source: own work).*

In the original *FedAVG* implementation by McMahan et al. (2016) the model is learned through stochastic gradient descent (SGD), where each party $k$ computes the average gradient $g_k = \nabla F_k(w_t)$ on its local data $n_k$ at the current model $w_t$ and iterates multiple times over the update $w_k \leftarrow w_k - \eta g_k$. The party submits the gradients to the central server, which aggregates the updates from all parties as:

$$w_{t+1} \leftarrow w_t - \sum_{k=1}^{K} \frac{n_k}{n} w_{t+1}^k$$

$$w_{t+1}^k \leftarrow w_t^k - \eta g_k, \forall k$$

While classic FedML operates on a client-server architecture, alternatives that do not rely on a central orchestrating server are also possible. For instance, parties can exchange model updates by establishing a peer-to-peer network, increasing the security of the process at the expense of consuming more bandwidth and resources for encryption (Roy et al., 2019).

Moreover, the distribution of features and samples across datasets may not be homogeneous. *Horizontal Federated Learning (HFL)* refers to the setup in which all datasets $\{D_{i=1}^K\}$ from the *K* parties contain different samples that share the same feature space. If instead, the same samples are present in all datasets, but feature spaces are disjoint, the setup is known as *Vertical Federated Learning (VFL)*.

Considering the high heterogeneity of data, especially if spread across different organizations, some authors have proposed to overcome the problem of sparse overlapping datasets through *Federated Transfer Learning (FTL)* (Liu et al., 2020). In this scheme, parties may select samples for training that minimizes the distance between their distributions (instance-based FTL) or learn a common feature space collaboratively (feature-based FTL). Alternatively, parties may start by using pre-trained models or by learning models from aligned samples to infer missing features and labels (model-based FTL).

Finally, it is important to note that the performances $v_{SUM}$ and $v_{FED}$ of the respective centralized and federated models, might differ considerably. This performance gap $\delta$ is characterized by $v_{SUM} - v_{FED} < \delta$ and will be strongly dependent on the characteristics of the particular application.

Consequently, FedML introduces a potential trade-off between the loss of performance respect to the centralized setup and the privacy guarantees provided by the distributed approach (Yang et al., 2019).

## 2.2 Collaborative Business Models

A business model describes essential aspects of an organization, explaining how the organization creates, delivers, and captures value (Osterwalder and Pigneur, 2010). In the academic literature, the definition of the term is fragmented, and no consistent boundaries are established. Nevertheless, it can be stated that a business model provides an organizational and strategic design for implementing a business opportunity (George, 2011).

In addition, Osterwalder and Pigneur (2010) argue that a shared understanding of the business model is crucial to its creation and success. Therefore, creating and discussing a business model requires a simple, relevant, and intuitively understandable concept without oversimplifying the complexity of how the organization works. The BMC by Osterwalder and Pigneur (2010) is a tool often used in practice to present a business model structured in nine components.

Business models are not only used for a single company but can also support assessing the feasibility and profitability of collaborations across companies (Kristensen and Ucler, 2016). The trend of an interconnected and dynamic environment encourages organizations to collaborate inter-organizationally and co-create value (Diirr and Cappelli, 2018). In literature, no unified framework exists for collaborations. Still, some approaches utilize Osterwalder and Pigneur (2010) general approach of a business model as a basis and customize it to set a higher focus on specifics (Kristensen and Ucler, 2016). For example, Eppinger and Kamprath (2011) highlight the importance of a partner and customer network in personalized medicine by modifying the canvas components and adding new ones, like intellectual property strategy. The approaches in the literature reach from modifications of business model components (e.g., Eppinger and Kamprath (2011) or Kristensen and Ucler (2016)), to configuration options of the business model (e.g., Curtis (2021) or Man and Luvison (2019)). However, the customizations are mainly application-oriented, tailored to the project to suit the needs and capture unique features influencing the business model and thus decisive for the project's success.

## 3 Methodology

This research was structured into three distinct parts. After a systematic literature review (SLR) to gain an overview of the challenges of collaborative business models, we organized an in-depth focus group interview to explore the novel field of inter-organizational FedML business models. By this, we aimed to augment the findings from the SLR and identify distinct challenges of business models for collaborative FedML projects. Since focus groups are characterized by their homogeneous group demographic, we pursued more generically applicable results by conducting additional semi-structured expert interviews. The research timeline is displayed in figure 2. The following subsections will go into more detail about the used research methodologies.

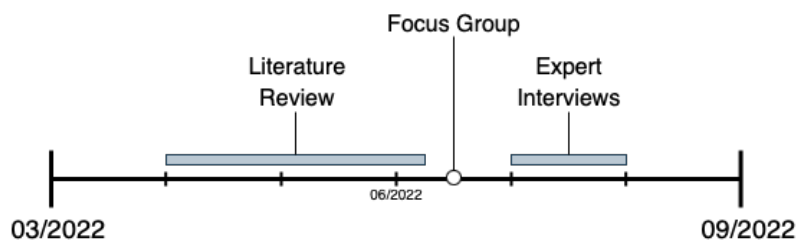

*Figure 2.* Research Timeline

## 3.1  Systematic Literature Review

To assess and identify the challenges of prevailing collaborative business models, we conducted a systematic literature review. By this, we intend to extract fundamentals and critical attributes of business models for inter-organizational collaborations, which will be collected, structured, and summarized. We followed a search strategy by Zhang et al. (2011) to identify the most relevant literature. Hence, the search is divided into base literature search, main search, and backward search.

The **Base Literature** consists of 13 papers including five publications on business model theory and eight on collaborative projects which were known by the authors prior to the search. Based on the initial literature corpus, we focused on finding keywords related to inter-organizational business models. The resulting search string *S* is as follows:

| String | Query |
|---|---|
| S1 | (collaborate* OR federated OR interorganization* OR inter-organization* OR intercompany OR cross-company OR multi-party OR cross-industr* OR multi-institution* OR shared OR sharing OR alliance OR networked) |
| S2 | ("business model*" OR "model canvas" OR "business value model*") |
| S | S1 AND S2 |

*Table 1.      Compiled search string for the database search.*

The **Main Search** was conducted from April 2022 to June 2022. The list of searched databases comprises *IEEE Xplore*, *ACM Digital Library*, *ScienceDirect, Wiley InterScience* and *SCOPUS*. We only included peer-reviewed English and German publications with full-text access. With the defined search string *S*, databases, and criteria we collected 262 distinct publications. We only aim to include work in the field of computer science and technology (coarse focus) as well as literature regarding inter-organizational collaborations and business models (narrow focus). Successively, the corpus consisting of 262 distinct publications has been filtered solely by title, abstract and full text regarding the defined coarse and narrow focus. By this, 18 publications remained.

For the **Backward Search**, we scanned the references of the resulting 18 publications from the main search. Again, these referenced publications were filtered by title, abstract and body according to the inclusion and exclusion criteria. After eliminating duplicates, we added one further study resulting in a total of 19 publications.

Finally, relevant information was extracted and synthesized from the final literature corpus. The structured and consolidated output yielded a set of critical attributes and challenges of business models for inter-organizational collaborations in the technological sector. In the following every insight from the SLR is referenced as usual with the corresponding publication.

## 3.2  In-depth Group Interview

Based on the identified challenges of prevailing collaborative business models from the SLR, we aimed to explore the distinct aspects of collaborative business models for inter-organizational FedML projects. Due to the novelty of the topic and the need for exploration, we organized an in-depth focus group interview (Dilshad and Latif, 2013) to study the business requirements of collaborative ML projects based on the findings from the SLR.

The focus group consisted of five participants and two moderators, where one moderator ensures smooth progress and the other ensures that all topics are covered. All participants worked on a project involving the adoption of FedML in a cross-company use case. The participants were briefed about collaborative business models and were given an overview of the findings from the SLR. Afterwards, the group was asked about the critical attributes and challenges of business models related to their collaborative FedML project followed by a reflection and lively discussion. Through this, we were able to identify further challenges based on their real-world experiences. The emerging data was coded by two researchers and incorporated into the results of the SLR. In the following, every insight which was gained through the in-depth focus group interview is referenced via the index *(FG)*.

### 3.3 Semi-Structured Expert Interviews

Even though the focus group helped explore the socio-technical challenges of FedML collaborations, the insights might be highly biased due to the homogeneous demographics of the participants. To gain a more generically applicable understanding, we aimed to draw from the experiences of further experts working in the field of applied AI, especially with experience in FedML projects. For these expert interviews, we draw from the Grounded Theory methodology (Hoda et al., 2011). Hence, we confronted the interviewees with a set of pre-defined questions and recorded as well as transcribed the interviews. We successively conducted and compared the results of each interview. After 5 interviews theoretical saturation was reached and consequently, the interview study was closed. The set of interviewees represented a more diverse set of experts from different organizations and domains. Table 1 presents a codified table of our sample. We developed an interview guide based on the research questions and findings from the SLR as well as the focus group interview including open questions about potentially missing attributes, challenges, and further insights. These interviews allowed us to go more in-depth and identify missing aspects and gain more detailed, in-depth individual understanding to develop the guideline questionnaire further.

The interviewees allowed the findings to be published in an anonymized manner but did not agree to disclose the full transcriptions. Therefore, the full transcripts are not included. The findings from the semi-structured interviews are referenced in the following with the participant ID as listed in table 2.

| **Participant ID** | **Position** | **Organization** | **Duration** |
|---|---|---|---|
| E1 | AI Business Developer | Large German software enterprise | 52 |
| E2 | AI Project Lead | Large German software enterprise | 44 |
| E3 | Principal Data Scientist | Large German software enterprise | 45 |
| E4 | Applied Researcher | Medium-sized innovation company | 35 |
| E5 | Scientific Researcher | Research institute for software development | 59 |

*Table 2.        Interview Study Participants*

## 4 Socio-Technical Challenges of Interorganizational Federated Machine Learning

Applying collaborative models can be challenging in different domains, especially when several companies are involved. When the business is operationalized, complexity increases significantly because the general business model idea needs to balance the interests of all participants (Pauna et al., 2021). Collaborations with multiple participants are complex in nature, and collaboration failure rates are high, leaving much revenue at risk and unrealized value (Man and Luvison, 2019). Moreover, aligning the business model with operational and governance-related aspects is suggested to help position the organization to deliver on its value proposition for a successful implementation of the business model (Curtis, 2021). Hence, early identification of the collaboration challenges is critical for the successful creation of the collaborative business model.

To better understand which specific collaboration challenges should be considered, we first investigate the challenges of prevailing inter-organizational business models and, secondly, which FedML-related socio-technical aspects are critical for successful implementation and therefore should be considered in a corresponding collaborative business model.

## 4.1   Challenges of Collaborative Business Models

Joint work of different organizations is complex, and organizations should be prepared to face challenges arising from cooperation. In the following, we give an overview of the systematized results of the SLR on the challenges of inter-organizational business models. We present our key results in a structured manner based on the work of Diirr and Cappelli (2018).

Diirr and Cappelli (2018) divide the challenges of inter-organizational collaborations into three categories: external, internal, and network-related challenges. External and internal challenges are detached from inter-organizational collaboration. External challenges relate to environmental challenges, e.g., natural events, and internal challenges arise from inside the project, for example, infrastructure problems. Network-related challenges focus on the relationships and interactions between organizations and can be further subdivided into management, business process, and collaboration challenges (Diirr and Cappelli, 2018). Based on these categories in conjunction with the findings of the SLR we derived the following network-related challenges as listed in table 3.

| Category | Description | Aspects |
|---|---|---|
| Management Challenges | Include how organizations create and establish collaboration, compromising the following aspects | Selection of suitable participating actors (Pauna et al., 2021). |
| | | Change management for dynamic collaboration (Caridà et al., 2015; Redlich et al., 2014). |
| | | Cooperation establishment: lack of commitment from participating organizations (Proulx and Gardoni, 2020); building and expanding trust between the parties (Bleja et al., 2020; Diirr and Cappelli, 2018; Redlich et al., 2014). |
| | | Decision-making and coordination slowness within the collaboration (Bleja et al., 2020; Caridà et al., 2015; Diirr and Cappelli, 2018; Redlich et al., 2014). |
| | | Communication with government authorities requires a different approach due to multiple parties' interactions (Pauna et al., 2021). |
| Business Process Challenges | Addresses the way organization's structure and design partnership operations | Definition of a mutual business goal of the collaboration (Diirr and Cappelli, 2018). |
| | | Co-Creation Management for delivering the value proposition:<br>• Distribution of financials, investment (Pauna et al., 2021), costs, and revenues (Bleja et al., 2020; Caridà et al., 2015; Pauna et al., 2021).<br>• Risk allocation (Diirr and Cappelli, 2018).<br>• Ownership structure (Diirr and Cappelli, 2018; Kujala et al., 2020).<br>• Responsibility assignment (Diirr and Cappelli, 2018).<br>• Align on quality of co-creation product (Diirr and Cappelli, 2018).<br>• Intellectual property Management (Eppinger and Kamprath, 2011). |
| | | Slower business strategy and process identification (Berkers et al., 2020; Diirr and Cappelli, 2018). |
| | | Alignment of the structures of heterogeneous organizations with distinct characteristics (Diirr and Cappelli, 2018). |
| | | Infrastructure for managing relationships between multiple collaboration actors (Caridà et al., 2015; Diirr and Cappelli, 2018). |
| Collaboration Challenges | Describe how organizations jointly work together to achieve the goal of collaboration | Agreement on collaboration and alignment with the organizations' own objectives (Bleja et al., 2020; Costa and Da Cunha, 2015; Diirr and Cappelli, 2018; Man and Luvison, 2019; Pauna et al., 2021). |
| | | Alignment of different organizations: culture and common ethics (Bleja et al., 2020; Diirr and Cappelli, 2018; Kujala et al., 2020). |
| | | Risk of opportunism of participants and consequences of action (Diirr and Cappelli, 2018). |

*Table 3.     Overview of Challenges for Inter-Organizational Collaborations*

This overview of challenges is a consolidation of the selected academic sources of the SLR and aims to provide a general understanding of the difficulties of such collaborations. It is important to note that naturally, this list might not be comprehensive and that certain, potentially important, aspects might be missing.

## 4.2   Aspects of Interorganizational FedML Business Models

To provide initial guidance in the creation of business models for collaborative FedML projects we aim to identify the corresponding critical challenges, which need to be considered at an early stage. These aspects are derived from literature research, in-depth group interviews and expert interviews. The following section presents the aspects in more detail. The basis is formed by the questions catalogue of the BMC by Osterwalder and Pigneur (2010). The extension reflects the specifics of collaboration and technology that can be generalized.

Before setting up a business model for inter-organizational FedML projects, it is necessary to clarify if the underlying problem can be solved by applying FedML. We presuppose a prior feasibility check and task-technology-fit analysis, but still include these two points in our systemized collection of socio-technical challenges.

Osterwalder and Pigneur (2010) use nine components with their associated questions to describe the generic business model clustered in three parts (Create Value, Deliver Value and Capture Value). We augmented these parts with a section that addresses specific aspects of the inter-organizational collaborative environment and FedML. The extensions were obtained from comparable business model extensions retrieved from the literature review, the challenges of section 4.1 and insights from the conducted interviews. Overall, this results in the list as seen in Table 4. The insights from the in-depth focus group interview are referenced by the index *FG* whereas the semi-structured interviews are referenced by the corresponding participant ID (E1 - E5) are described in Table 2.

The complete list with guiding questions including the used literature corpus is provided in a complementary document[2]. The overview of aspects targets important areas of a business model for collaborative AI projects, but due to the nature of the research approach, this list might not be exhaustive.

---

[2] Extended Business Model Canvas, Guiding Questions and Literature Corpus: bit.ly/3mXOUQg

| Section | Component | Description |
|---|---|---|
| Create Value | Value Proposition | Describes the value that can be delivered to a certain customer. |
| | Customer Relationships | Explains the kinds of relationships an organization makes with particular customer segments. |
| | Channels | Defines how an organization interacts and reaches its customers to serve the value proposition. |
| | Customer Segments | Refers to the various groups of people the organization wants to reach and serve. |
| Deliver Value | Key Partners | Designates the network of suppliers and partners that are essential business model. |
| | Key Activities | Describes the most important things an organization must do to make its business model work. |
| | Key Resources | Defines the essential resources required. |
| Capture Value | Cost Structure | Describes all costs that are decisive for the operation of the business model |
| | Revenue Streams | Represent the earnings that an organization receives from each customer segment. |
| Collaboration Management | Collaboration Structure | Describes the negotiation mechanisms for building the network of participants and how decision-making is handled and coordinated within the collaboration. This mainly reflects whether there is a dominant participant in the collaboration (FG, E1, E2, E3, E4). |
| | Participant Management | Includes the formation regarding the suitability of participants, the change management of the collaboration, and the transparency of the project participants (FG, E1, E4). |
| | Infrastructure | Includes the management of collaboration- and technology-specific communication channels, as well as the platform to facilitate it (E1, E4). |
| Co-Creation Management | Distribution of Ownership, Responsibility and Accountability | Encourages distribution mechanisms within the collaboration. The aspects mentioned are crucial for the product created by the ML model (FG, E1, E2, E3, E4, E5). |
| | Distribution of Revenue and Costs | Describes how participants are rewarded for their participation and how additional effort is compensated (FG, E1, E2, E3, E4). |
| | Intellectual Property Management | Is crucial for enabling inter-organizational collaboration for joint activities (FG, E1, E2, E3, E4). |
| Co-Creation Practices | Profit Calculation | Describes the project's estimated cost-effectiveness over the FedML lifecycle with the different participants (E2, E3, E4). |
| | Risks of Infeasibility | Specify how the feasibility of solving the project with the FedML technology is determined (FG, E1, E4). |
| | Alignment in Quality | Describes how product quality is defined, ensured, and tracked throughout the FedML lifecycle (E1, E5). |
| | Implementation of Activities | Explains the extent to which the data-generating parties are involved in operational implementation (FG, E4, E5). |
| FedML Product | Compliance Data Protections | Include what regulations must be considered to develop a compliant FedML model (FG, E1, E4). |
| | Versioning | How versioning is handled within the FedML process (E1). |
| | Retirement | Includes how an ML Model can be recalled and claimed from participating parties and also end customers (E1, E4). |

*Table 4.     Aspects of Inter-Organizational FedML Collaborations*

## 4.3    Extension of the Business Model Canvas

The BMC by Osterwalder and Pigneur (2010) can be used to guide the creation and discussion of a business model. The canvas is a simplification of reality and should also pick up the most critical aspects of the business model already. To provide an easily consumable entry point for the initial viability assessment of collaborative AI projects, we aimed to extend the traditional BMC with the corresponding most critical aspects and challenges. This extended BMC represents another layer of simplification to the provided collection of socio-technical challenges from chapter 4.2. Hence, the presented collection was reduced to the most critical subareas: *Collaboration Management* and *Co-Creation Management*. Both aspects were mentioned as the most critical challenges in the interviews and address the collaborative approach as well as the joint creation of a FedML model. Guiding questions are added to the component titles for ease of use and intuitive comprehension. Figure 3 shows the extension of the canvas with colour-coded support. White tiles represent the business model aspects of a collaboration as a whole and blue tiles the aspects within the collaboration. To spare time and efforts, we suggest that the extended BMC should be used as a first basis to identify potential roadblocks of collaborative FedML projects. Thereupon, decision-makers can use the more detailed and comprehensive collection of socio-technical challenges as a second step of the viability assessment. To develop a concrete collaborative business model further steps (e.g. the identification of value streams between the collaborating parties) are necessary. Our artefacts solely represent a one-stop shop for the early identification of socio-technical aspects, challenges and potential roadblocks in the creation of collaborative AI projects.

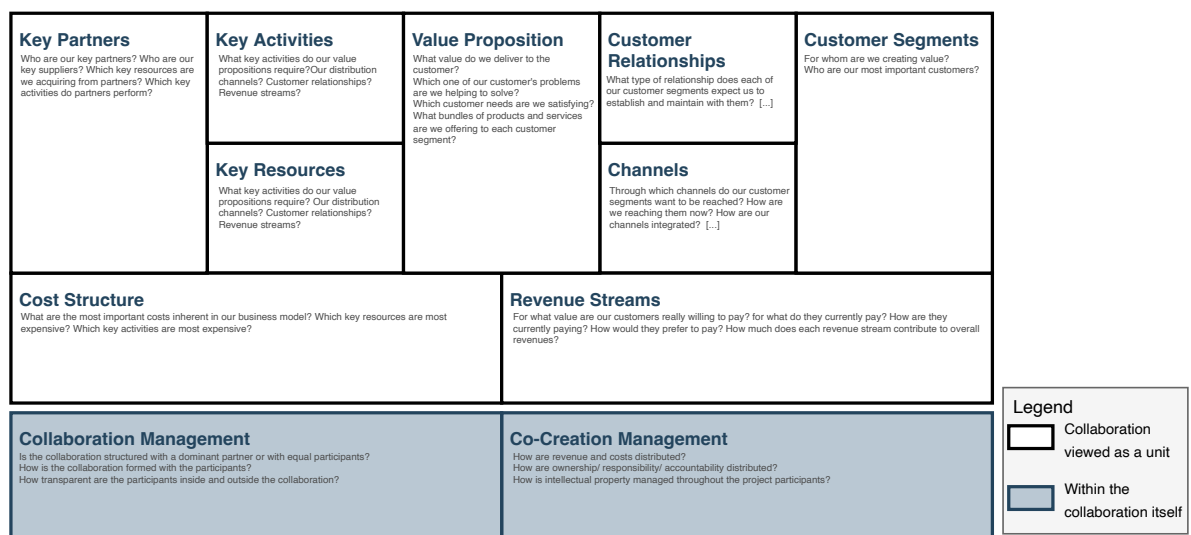

*Figure 3.    Extended BMC for Inter-Organizational FedML Collaborations*

## 5    Conclusion

A significant amount of the world's data is scattered, stored, and locked up in decentralized data silos. The siloed data is hardly accessible, which leaves a large portion of already generated data and its economic potential largely untapped. The model-to-data approach of FedML technically enables the creation of a joint ML model from decentralized data without the need of data sharing. However, this novel privacy-enhancing ML paradigm requires collaboration of multiple parties which own the data silos. Consequently, a collaborative business model is required to define how value can be created. Setting up these collaborative business models is complex with a high potential of failure.

The information systems research literature offers interesting insights on emerging AI business models and collaborative business models. Current research on the distinct aspects of AI business models resulted in a multitude of relevant contributions. For example, how organizations evaluate AI value creation mechanisms (Alsheiabni et al., 2020) or how the value drivers of AI open new business model design opportunities (Böttcher et al., 2022). Additionally, studies investigated how AI already transformed the business models of specific domains, such as the insurance sector (Zarifis et al., 2019; Zarifis and Cheng, 2023), Fintech (Zarifis and Cheng, 2022) or education (Renz and Hilbig, 2020; Zarifis and Efthymiou, 2022). Our study complements the current information systems literature by revealing the socio-technical challenges of collaborative AI projects and consequently providing guidance in the creation of a corresponding inter-organizational business model.

## 5.1 Contributions

Through a systematic literature review, in-depth focus group interview and semi-structured expert interviews, we first investigated the challenges and aspects of business models for inter-organizational FedML projects. These findings were aggregated and structured into a comprehensive set of guiding questions and compressed into an extension of the BMC. The resulted questionnaire represents a set of detailed aspects which need to be considered in the creation of the collaborative business model. Thereby, we aid decision-makers at an early stage of the business model development and prepare them for challenges related to the collaboration. The traditional BMC by Osterwalder and Pigneur (2010) was complemented by two dimensions and should aid decision-makers in the first assessment of value creation, delivery and capturing for inter-organizational FedML projects. We can sum up the results of our research questions as follows:

**RQ1:** *What are the general challenges of collaborative business models?*

A joint work of multiple organizations is complex, and a multitude of challenges arise from cooperation. Through a systematic literature review on collaborative business models in the technology sector, we aggregated a list of said challenges. Based on the example of Diirr and Cappelli (2018), we structured this list into *Management Challenges*, *Business Process Challenges*, and *Collaboration Challenges*. These categories were filled with challenges from the identified literature corpus. Specifically, the cooperation establishment, slowness within the collaboration, distribution of financials, and agreement on collaboration and alignment with the organizations' own objective seem to be the most critical challenges. However, this list may not be exhaustive but should present the most important challenges.

**RQ2:** *What are* the *aspects of inter-organizational FedML business models in relation to prevailing collaborative business models?*

To identify relevant aspects of inter-organizational FedML business models, we organized an in-depth focus group interview to explore this novel topic followed by semi-structured expert interviews to get a more diverse view and to augment the findings from the focus group study. Particularly, the difficulties of allocating rights and responsibilities within the co-creation management appear to be of specific interest in the inter-organizational use of FedML. These results were combined with the identified challenges from RQ1 and structured into groups of aspects for collaborative FedML business models. As a result, we received four aspect clusters: *Collaboration Management*, *Co-Creation Management*, *Co-Creation Practices*, and *FedML Product*. A more detailed specification is listed in the provided collection of socio-technical challenges.

**RQ3:** *Which aspects and attributes should be considered for inter-organizational FedML projects and how can these be structured into an extended BMC?*

The findings from RQ2 and the resulting questionnaire act as a basis for the extended BMC. Since the canvas should be a simplification of reality but should also consider the most critical aspects of the business model, we selected the most referenced and mentioned subareas. Therefore, *Collaboration Management* and *Co-Creation Management* were selected to expand the original canvas. These subareas seemed to be the most critical challenges of prevailing collaborative business models and were mentioned as the most critical success factors in the focus group interview as well as in the expert interviews. Hence, we argue that these two dimensions are important extensions to the BMC and capture the most critical aspects without losing handiness. This assumption and reasoning needs to be validated in practice but should provide a proper basis.

Overall, the insights of our study advance the understanding of the socio-technical challenges which arise in collaborative AI projects and of the relevant aspects in the development of corresponding inter-organizational business models. The systemized list of socio-technical challenges and resulting extended BMC can be used by decision-makers for the initial viability assessment. The comprehensive set of guiding questions can be used as assistance for the business model development process. By this, we guide decision-makers to make use of FedML and provide support to overcome socio-technical obstacles to the adoption of collaborative AI. Therefore, our study helps to unlock previously inaccessible value from siloed data and contributes to business model innovation.

## 5.2 Limitations

There are obvious limitations to our work. The participants of the focus group were affiliated with the same company and worked on similar projects within this company. Hence, the findings from this group could be highly biased and might have led to one-sided results. We tried to counteract this by conducting further interviews with experts of different backgrounds and affiliations. However, the theoretical saturation was reached after five interviews which terminated our interview study with a small sample size of five participants. More data and consequently more interesting perspectives from a bigger and even more diverse set of interviewees might enrich our results. Therefore, we encourage researchers and generally interested readers to use our work as a basis to complement, refine and develop our artefacts with their own insights. Moreover, our work poses as a starting point for the development of potential business models of inter-organizational FedML projects. This model only comprises the relevant decisive factors for the success of collaborative FedML projects but does not capture more low-level aspects as the value streams between organizations. This would pose a natural next step in developing a potential business model. We encourage the investigation of a model which captures how actors within the collaborative FedML project might exchange value (e.g., e3-value model). We also assume that a feasibility check was conducted beforehand if FedML is a fitting solution. If FedML can be excluded from the set of reasonable technologies choices for the given problem, there would be no point in going a step further by addressing the multitude of distinct socio-technical challenges of collaborative AI projects and building a concrete business model. Therefore, an a priori task technology analysis would be reasonable.

## 5.3 Future Research

Generally, the research in the field of FedML is dominated by technical work. Nonetheless, the practical adoption of novel technologies like FedML is dependent on more than the technical dimension. Decision-makers will not consider using FedML if the legal framework is fuzzy, the business model does not provide a proper value proposition or the task and technology contradict. We believe that FedML is a technology with great potential and will open a large variety of possibilities in the era of big data, where huge amounts of data is stored in data silos. To unlock this potential, there needs to be more research on the social and socio-economic challenges of collaborative ML. From legal frameworks and governance concepts to task-technology analyses, the research field is wide open and ready to be explored.

# Acknowledgments

The authors would like to thank SAP SE for supporting this work.

# References

Alsheibani, S. A., Cheung, Y., Messom, C., and Alhosni, M. (2020). "Winning AI Strategy: Six-Steps to Create Value from Artificial Intelligence," in: Anderson, B. B., Thatcher, J., Meservy, R. D.,

Chudoba, K., Fadel, K. J., Brown, S. (eds.) *26th Americas Conference on Information Systems,* Virtual.

Berkers, F., Turetken, O., Ozkan, B., Wilbik, A., Adali, O. E., Gilsing, R., and Grefen, P. (2020). "Deriving Collaborative Business Model Design Requirements from a Digital Platform Business Strategy," *IFIP Advances in Information and Communication Technology* 598 (1), 47–60.

Bleja, J., Wiewelhove, D., Grossmann, U., and Mörz, E. (2020). "Collaborative Business Model Structures for Wireless Ambient Assisted Living Systems," *2020 IEEE 5th International Symposium on Smart and Wireless Systems within the Conferences on Intelligent Data Acquisition and Advanced Computing Systems (IDAACS-SWS)*, Virtual.

Böttcher, T., Weber, M., Weking, J., Hein, A., and Krcmar, H. (2022). "Value Drivers of Artificial Intelligence". *28th Americas Conference on Information Systems,* Minneapolis, USA.

Caridà, A., Colurcio, M., and Melia, M. (2015). "Designing a collaborative business model for SMEs," *Sinergie Italian Journal of Management* 33 (1), 233–253.

Costa, C. C., and Da Cunha, P. R. (2015). "More than a gut feeling: Ensuring your inter-organizational business model works," *28th Bled EConference: #eWellbeing - Proceedings*, 86–99.

Curtis, S. K. (2021). "Business model patterns in the sharing economy," *Sustainable Production and Consumption* 27, 1650–1671.

Diirr, B., and Cappelli, C. (2018). "A systematic literature review to understand cross-organizational relationship management and collaboration." *51st Hawaii International Conference on System Sciences,* Hawaii, USA.

Dilshad, R. M., and Latif, M. I. (2013). "Focus Group Interview as a Tool for Qualitative Research: An Analysis," *Pakistan Journal of Social* Sciences 33, 191-198.

Eppinger, E., and Kamprath, M. (2011). "Sustainable Business Model Innovation in Personalized Medicine," *R&D Management Conference,* Linköping, Sweden.

Forradellas, R. F. R., and Gallastegui, L. M. G. (2021). "Digital Transformation and Artificial Intelligence Applied to Business: Legal Regulations, Economic Impact and Perspective," *Laws*, 10 (3), 70.

George, G. (2011). "The business model in practice and its implications for entrepreneurship research. Entrepreneurship Theory and Practice," *Entrepreneurship Theory and Practice* 35 (1), 83-111.

Hoda, R., Noble, J., and Marshall, S. (2011). "Grounded theory for geeks," *Proceedings of the 18th Conference on Pattern Languages of Programs - PLoP '11*, Irsee, Germany.

Kristensen, K., and Ucler, C. (2016). "Collaboration Model Canvas: Using the Business Model Canvas to Model Productive Collaborative Behavior," *2016 International Conference on Engineering, Technology and Innovation/IEEE International Technology Management Conference (ICE/ITMC)*, Wuhan, China.

Kujala, J., Aaltonen, K., Gotcheva, N., and Lahdenperä, P. (2020). "Dimensions of governance in interorganizational project networks," *International Journal of Managing Projects in Business* 14 (3), 625–651.

Li, J., Zhang, C., Zhao, Y., Qiu, W., Chen, Q., and Zhang, X. (2022). "Federated learning-based short-term building energy consumption prediction method for solving the data silos problem," *Building Simulation* 15 (6), 1145–1159.

Liu, Y., Kang, Y., Xing, C., Chen, T., and Yang, Q. (2020). "A Secure Federated Transfer Learning Framework," *IEEE Intelligent Systems* 35 (4), 70–82.

Man, A.-P. de, and Luvison, D. (2019). "Collaborative business models: Aligning and operationalizing alliances," *Business Horizons* 62 (4), 473–482.

McMahan, H. B., Moore, E., Ramage, D., and Arcas, B. A. y. (2016). "Federated Learning of Deep Networks using Model Averaging," *CoRR*.

Osterwalder, A., Pigneur, Y., and Clark, T. (2010). *Business model generation: A handbook for visionaries, game changers, and challengers*, Wiley.

Pauna, T., Lampela, H., Aaltonen, K., and Kujala, J. (2021). "Challenges for implementing collaborative practices in industrial engineering projects," *Project Leadership and Society 2*, 100029.

Proulx, M., and Gardoni, M. (2020). "Methodology for Designing a Collaborative Business Model – Case Study Aerospace Cluster," *IFIP Advances in Information and Communication Technology* 594, 387–401.

Redlich, T., Basmer, S.-V., Buxbaum-Conradi, S., Krenz, P., Wulfsberg, J., and Bruhns, F.-L. (2014). "Openness and trust in value co-creation: Inter-organizational knowledge transfer and new business models," In P. G. Kocaoglu D.F. Anderson T. R.,. Daim T. U. ,. Kozanoglu D. C. ,. Niwa K. (Ed.), *PICMET 2014—Portland International Center for Management of Engineering and Technology, Proceedings: Infrastructure and Service Integration*, Portland, USA.

Renz, A., and Hilbig, R. (2020). "Prerequisites for artificial intelligence in further education: identification of drivers, barriers, and business models of educational technology companies," *Int J Educ Technol High Educ*, 17(14).

Roy, A. G., Siddiqui, S., Pölsterl, S., Navab, N., and Wachinger, C. (2019). "BrainTorrent: A Peer-to-Peer Environment for Decentralized Federated Learning," *arXiv*.

Schomakers, E.-M., Lidynia, C., and Ziefle, M. (2020). "All of me? Users' preferences for privacy-preserving data markets and the importance of anonymity," *Electronic Markets* 30 (3), 649-665.

Yang, Q., Liu, Y., Cheng, Y., Kang, Y., Chen, T., and Yu, H. (2019). "Federated learning," *Synthesis Lectures on Artificial Intelligence and Machine Learning* 13 (3), 1–207.

Zarifis, A., and Cheng, X. (2022). "A Model of Trust in Fintech and Trust in Insurtech: How Artificial Intelligence and the Context Influence It," *Journal of Behavioral and Experimental Finance* 36 (1), 1-20.

Zarifis, A., and Cheng, X. (2023). "AI is Transforming Insurance With Five Emerging Business Models," *Encyclopedia of Data Science and Machine Learning.* IGI Global, 2086-2100.

Zarifis, A., and Efthymiou, L. (2022). "The four business models for AI adoption in education: Giving leaders a destination for the digital transformation journey," *2022 IEEE Global Engineering Education Conference (EDUCON)*, Tunis, Tunisia.

Zarifis, A., Holland, C. P., and Milne, A. (2019). "Evaluating the Impact of AI on Insurance: The Four Emerging AI and Data Diven Business Models," *Emerald Open Research*, 1-17.

Zhang, H., Babar, M. A., and Tell, P. (2011). "Identifying relevant studies in software engineering," *Information and Software Technology 53* (6), 625–637.